\documentclass[twoside,11pt]{article}

\usepackage{jmlr2e}
\usepackage{amsmath,amssymb,amsfonts}
\usepackage{algorithm}
\usepackage{algpseudocode}
\usepackage{graphicx}
\usepackage{textcomp}
\usepackage[usenames,dvipsnames]{xcolor}
\usepackage{hyperref}
\usepackage{float}
\usepackage{svg}
\usepackage{tabularx}
\usepackage{booktabs}       
\usepackage{mathtools}
\usepackage{siunitx}
\usepackage{url}
\usepackage{subfigure}
\usepackage{caption}
\usepackage{subcaption}

\usepackage{lastpage}

\ShortHeadings{Measuring the Energy Consumption and Efficiency of Deep Neural Networks}{Tripp, Perr-Sauer, Gafur, Nag, Purkayastha, Zisman, and Bensen}

\firstpageno{1}

\title{Measuring the Energy Consumption and Efficiency of Deep Neural Networks: An Empirical Analysis and Design Recommendations}

\begin{document}

\newcommand{\authctripp}{Charles~Edison~Tripp}
\newcommand{\authjperrsauer}{Jordan~Perr-Sauer}
\newcommand{\authjgafur}{Jamil~Gafur}
\newcommand{\authebensen}{Erik~A.~Bensen}
\newcommand{\authanag}{Ambarish~Nag}
\newcommand{\authavip}{Avi~Purkayastha}
\newcommand{\authszisman}{Sagi~Zisman}




\author{\name Charles Edison Tripp \email Charles.Tripp@nrel.gov \\
        \name Jordan Perr-Sauer \email Jordan.Perr-Sauer@nrel.gov \\
        \name Jamil Gafur \email Jamil.Gafur@nrel.gov \\
        \name Ambarish Nag \email Ambarish.Nag@nrel.gov \\
        \name Avi Purkayastha \email Avi.Purkayastha@nrel.gov \\
        \name Sagi Zisman \email Sagi.Zisman@nrel.gov \\
        \addr Computational Science Center\\
        National Renewable Energy Laboratory\\
        Golden, CO 80401, USA
        \AND
        \name Erik A. Bensen \email ebensen@andrew.cmu.edu \\
        \addr Department of Statistics and Data Science\\
        Carnegie Mellon University\\
        5000 Forbes Ave\\
        Pittsburgh, PA 15213, USA
}

\editor{}



\maketitle

\begin{abstract}
Addressing the so-called ``Red-AI'' trend of rising energy consumption by large-scale neural networks, this study investigates the actual energy consumption, as measured by node-level watt-meters, of training various fully connected neural network architectures. 
We introduce the BUTTER-E dataset, an augmentation to the BUTTER Empirical Deep Learning dataset, containing energy consumption and performance data from 63,527 individual experimental runs spanning 30,582 distinct configurations: 13 datasets, 20 sizes (number of trainable parameters), 8 network ``shapes'', and 14 depths on both CPU and GPU hardware collected using node-level watt-meters.
This dataset reveals the complex relationship between dataset size, network structure, and energy use, and highlights the impact of cache effects. 
We propose a straightforward and effective energy model that accounts for network size, computing, and memory hierarchy. 
Our analysis also uncovers a surprising, hardware-mediated non-linear relationship between energy efficiency and network design, challenging the assumption that reducing the number of parameters or FLOPs is the best way to achieve greater energy efficiency. 
Highlighting the need for cache-considerate algorithm development, we suggest a combined approach to energy efficient network, algorithm, and hardware design. 
This work contributes to the fields of sustainable computing and Green AI, offering practical guidance for creating more energy-efficient neural networks and promoting sustainable AI.
\end{abstract}
\begin{keywords}
Green AI, Red AI, Sustainable Computing, Neural Networks, Deep Learning, Energy Efficiency, Energy Measurement, Dataset, Energy Model
\end{keywords}

\section{Introduction}
\label{sec:introduction}
Large High-Performance Computing (HPC) systems and their supporting infrastructure constitute a significant and growing portion of electrical energy consumption. 
One projection forecasts a substantial 20.9\% proportion of the world's total electricity demand will go to computing by 2030 \cite{jones2018stop}.
Despite significant advancements in chip technology, driven by Moore's Law and continual improvement in energy efficiency~\cite{andrae_new_2020,Moore1965, dennard1974design,Dennard1999,Koomey2011,Markov2014}, the ever-growing demand for computational power and data storage has outpaced these gains. This trend is evident in the increasing energy consumption of ever-more powerful computing systems~\cite{Top500}.
\begin{figure}[p]
    \centering
    \includegraphics[width=\linewidth]{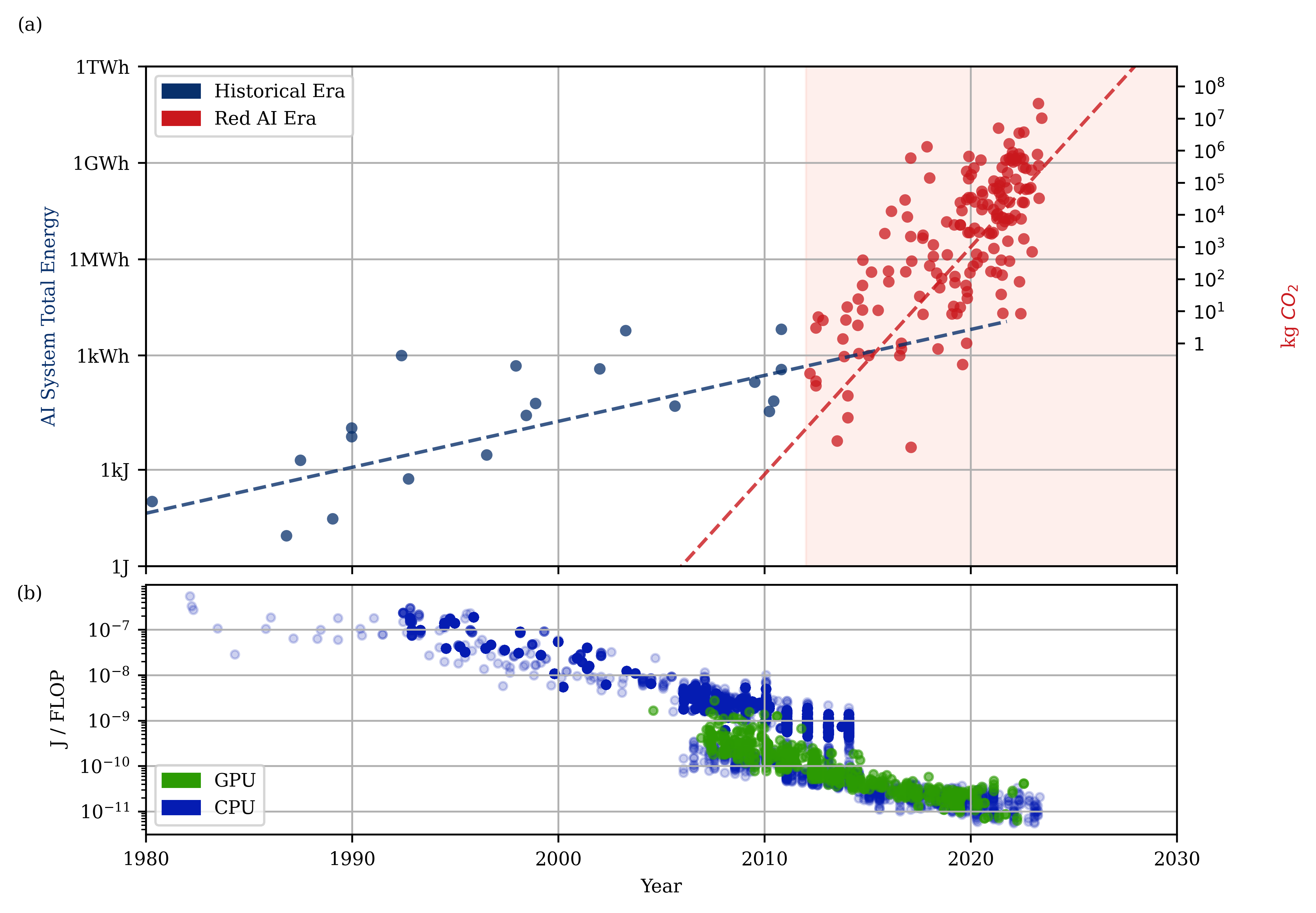}
    \caption{(a) The energy consumption (left axis) and corresponding carbon emissions given the average energy generation mix in the United States (right axis) incurred by training published AI models has increased dramatically over the last two decades, the so-called ``Red AI Era.''
    The AI System Total Energy is an computed metric, which accounts for efficiency gains in hardware (shown in (b)) and data center PUE.
    Note the logarithmic scale on the vertical axes for both (a) and (b).
    To compile this data we used CPU~DB~\cite{danowitz2012cpu}, Intel, AMD, and NVIDIA manufacturer websites, and compiled tables on Wikipedia to estimate J/FLOP using thermal design power (TDP) and published throughput (FLOP/s) or estimated it using instruction set, clock frequency, and core count.
    We incorporated improvements in PUE~\cite{statista_2023} and used the United States national average CO$_2$ emissions per kWh in 2021~\cite{eGRID_data_explorer_2024}. The number of parameters/FLOPs for each model are from ~\cite{epochMachineLearningData2023}.
    }
    \label{fig:trends:ai_energy_co2_cpu_gpu}
\end{figure}
Over the last decade, Artificial Intelligence (AI) has emerged as a dominant and energy-intense application, with energy usage growing exponentially, far outpacing hardware efficiency gains.
Without substantial advances in ``GreenAI'' technologies to counter this ``RedAI'' trend, we are  on course to dive head-first into the troubling and uncharted waters of unprecedented high levels of computational energy and carbon emissions~\cite{schwartz2020green,Strubell2019,Rolnick2019TacklingCC}.
Some may challenge this statement on the basis that computing hardware will become increasingly efficient, obviating the need for more efficient AI systems.
Others may question if improvements in Power Usage Effectiveness (PUE) will mitigate rising datacenter and High-Performance Computing (HPC) AI energy consumption.
And, others may object that rising renewable and low-carbon intensity generation will obviate the need for energy efficient AI.
However, even when accounting for each of these three mitigating factors in Figure \ref{fig:trends:ai_energy_co2_cpu_gpu}: increasing hardware energy efficiency, improvements in PUE, and decreases in CO$_2$ emissions per kWh of electricity; the training energy of state-of-the-art AI models continue to double in both CO$_2$ emissions and energy consumption every four to six months in this ``RedAI'' era.
And these exponential trends are multiplied by the ongoing rapid proliferation of AI systems not captured in this figure, such as the widespread commercial use of model inference, and the increasing number of organizations and individuals training these models.
In response to the swift rise in energy used for deep learning, this work includes the following contributions:
\begin{itemize}
    \item We introduce the BUTTER-E dataset, available for free through the OpenEI Data Portal,\footnote{The BUTTER-E dataset is available at: \url{https://data.openei.org/submissions/5991}} measuring the real-world energy consumption of training dense fully-connected neural networks in an HPC datacenter containing energy consumption and performance data from 63,527 individual experimental runs spanning 30,582 distinct configurations: 13 datasets, 20 sizes (number of trainable parameters), 8 network ``shapes'', and 14 depths on both CPU and GPU hardware collected using node-level watt-meters.
    \item We characterize non-linear hardware-mediated energy-hyperparameter interactions such as between parameter count and energy cost, and simpler ones including the linear relationship between training set size and energy per epoch. The code to reproduce these analyses is available for free through Github.\footnote{The code to reproduce the analysis is available at \url{https://github.com/NREL/BUTTER-E-Empirical-analysis-of-energy-trends-in-neural-networks-supplementary-code}}
    \item Using these interactions, we propose and fit a simple but effective energy model of the energy consumption of fully connected neural networks.
    \item Merging energy measurements with the BUTTER dataset, \cite{butter_oedi_osti_dataset}, we characterize the impact of hyperparameter choice on energy efficiency across all training tasks and architectures, and identify some counter-intuitive results.
    \item We use these results to motivate and propose research directions in energy-efficient architectures, algorithms, and hardware to meet the rising challenge of ``Red AI''.
\end{itemize}

The rest of this paper is structured as follows:
Section \ref{sec:related_work} discusses related work.
Section \ref{sec:the_dataset} describes the BUTTER-E dataset and method used to collect and compile the data.
Section \ref{sec:data_quality} describes how the data was prepared for the analysis in this paper, presenting practical challenges and the techniques we used to standardize and compare real-world computational energy measurements in a large HPC system.
Section \ref{sec:aggregate_trends} discusses observed aggregate relationships between hyperparameter (and dataset) choice and energy consumption.
Section \ref{sec:cache_effects} provides a detailed analysis of the interaction between cache size and energy consumption by linking network topology defined factors to critical working sets and mapping those working sets onto the memory hierarchy.
Section \ref{sec:energy_model} uses the results from Sections \ref{sec:aggregate_trends} and \ref{sec:cache_effects} to propose a hardware-informed energy model describing the salient relationships between network topology and energy consumption.
Section \ref{sec:energy_efficiency} joins the BUTTER-E and BUTTER datasets to inspect the surprising key relationship we observed between hyperparameter choice, model performance, and energy efficiency.
Finally, Section \ref{sec:conclusion} proposes future work informed by this study to optimize hardware and software for energy efficient deep learning.

\section{Related Work}
\label{sec:related_work}

The investigation of Neural Network (NN) Architecture on various AI applications and its impact on performance and energy efficiency has garnered  significant attention in the research community. 
As the complexity and diversity of real-world applications that can be solved with neural networks continue to expand, researchers have studied various architectural designs to enhance their performance. 
For instance, \cite{moon2019comparative} established through experimental results that ``the type of activation function and the number of hidden layers are critical in the performance of neural networks.''
Furthermore, a recent survey, \cite{kaviani2020influence}, indicates that artificial neural networks characterized by ``alternative complex topologies, rather than being fully connected, show high performance with less complexity'', computational time, and energy consumption. 

This prompts a pertinent question: does the exponential surge in number of parameters translate into a commensurate exponential upswing in energy consumption? 
\cite{desislavov2021compute} endeavors to address this query, with a specific focus on inference costs, as these costs predominate due to larger matrix-matrix multiplication.
As problem complexities increase, model developers have scaled up model architectures to handle these problems.
By increasing the parameter counts, topological complexity, and quantity of training data, developers aim to model increasingly detailed relationships.
While this brute-force approach has led to improved accuracy, performance, and capability, these gains have come at a cost.
More complex models use more energy to train and apply, resulting in higher monetary costs, greater carbon emissions, and in many cases, longer development, training, and inference times.

The field of Neural Architecture Search (NAS) concerns the study and optimization of the hyperparameters that define neural network architectures.
Benchmark datasets, such as NAS-Bench 101, \cite{ying2019bench}, evaluate many of NN architectures and their associated loss and accuracy scores after training.
In EC-NAS, \cite{bakhtiarifard_ec-nas_2023} estimates the energy consumption and carbon emissions of each architecture in NAS-Bench 101. 
In that work, the authors re-trained 1\%  the NAS-Bench architectures, measuring the energy consumption using on-chip hardware performance counters. 
A multi-layer perceptron model was then trained to interpolate this subset to the entire dataset resulting in a non-analytic surrogate energy model.

A number of different studies have looked at the environmental impacts of various machine learning architectures, strategies and types of input features. 
\cite{parcollet_energy_2021} found that 50\% of the energy consumed in training modern large language models (LMMs) is spent to achieve the final 0.3\% improvement in accuracy.
This result was corroborated in another study on multi-layer perceptrons, \cite{brownlee_exploring_2021}, where the authors find that ``30-50\%'' of the energy consumed in training can be eliminated with only minimal loss in accuracy. 
\cite{garcia2019estimation} reviewed computational energy consumption estimation approaches and mapped these approaches to ML applications via an extensive literature survey.
They found that there are two research directions to AI energy prediction: energy prediction modeling as seen in NeuralPower~\cite{cai2017neuralpower}, and integrated power monitoring tools such as in SyNERGY, \cite{rodrigues2018synergy}.

\cite{xu2023energy} estimate the environmental impact of Deep Convolutional NN model architectures in terms of the energy consumed and CO$_2$ emissions produced during training.
This is the only other study the authors are aware of that, like the BUTTER-E dataset, measured real-world, node-level power with a wattmeter. 
The vast majority of the other studies rely on hardware performance counters or more abstract (and many times not empirically validated) models, such as simple estimates of FLOPs, \cite{liu2021freetickets},or Transistor Operations, \cite{li_transistor_2022}, to provide an estimate of the energy consumption.
While these models are a step towards grappling with the energy implications of deep learning, anecdotally our experience suggests that these models may not capture the nuances of the relationship and tend to underestimate the total energy consumed due to their focus on the energy expenditure of subsystems (CPU, RAM), and therefore could lead researchers and practitioners to false conclusions.
Impressively, the proposed model of CNN energy consumption in \cite{yang_method_2017}, while not empirically substantiated, contains many similar elements to the model we propose.
In particular, Yang identifies the access patterns and cache interactions of the critical working sets as a key step in understanding the energy costs of deep learning workloads, a conjecture that is empirically substantiated by this study.

The urgent need to address AI's energy efficiency has also been raised by the wider computer science community. 
In recognition of the importance of energy-efficient AI, the IEEE organization has taken a proactive step by introducing the ``Low-Power Image Recognition Challenge.'' \footnote{https://lpcv.ai}
This challenge aims to incentivize researchers and practitioners to develop innovative solutions that minimize energy consumption while maintaining or even improving the accuracy of image classification tasks. 
By encouraging the design of energy-efficient neural network architectures, the challenge seeks to address the pressing need for sustainable and environmentally friendly machine learning practices.

\section{Introducing the Butter-E Dataset}
\label{sec:the_dataset}

Here, we present the BUTTER-E data-set, which provides real-world energy measurements of multi-layer perceptron (MLP) training experiments spanning 41,129 distinct hyperparameter combinations: 13 datasets, $2^{5}$ to $2^{25}$ parameters distributed according to 8 distinct network ``shapes" across 2-20 layers, on both CPU and GPU hardware.
Training datasets were drawn from the Penn Machine Learning Benchmark database: 201\_pol, 294\_satellite\_image, 529\_pollen, 537\_houses, adult, banana, connect\_4, MNIST, nursery, sleep, splice, and wine\_quality\_white \cite{Olson2017PMLB, romano2021pmlb, deng2012mnist}.
The number of trainable parameters (NTP) was swept across $\{2^n \forall n \in \{5,6,..., 25\}\}$.
Network shapes tested were rectangle (all hidden layers have equal width), rectangle\_residual (rectangular with residual connections), trapezoid (widths linearly decrease with depth), exponential (widths decrease exponentially from input to output), and wide\_first\_[n]x for $n \in \{2,4,8,16\}$ (the first hidden layer is $N$ times wider than the deeper hidden layers).
The depths tested were $\{2, 3, 4, 5, 6, 7, 8, 9, 10, 12, 14, 16, 18, 20\}$. 
More information about these hyperparameters and experimental configuration can be found in the BUTTER dataset and its associated documentation, \cite{butter_publication}.
A batch size of $256$ was used along with a learning rate of $0.001$ setting of the Adam optimizer, \cite{KingBa15}.
Experiments used Tensorflow and were configured to use Intel's OneDNN when running on CPUs and cuDNN (NVIDIA CUDA Deep Neural Network) on GPUs.

BUTTER-E enables researchers and practitioners to study how energy consumption is affected by the topology of the network and the trade-offs between energy consumed, epochs trained, and loss level achieved.
BUTTER-E is intended to be joined with the BUTTER dataset which characterizes the performance of 483k distinct fully connected neural networks but does not include energy measurements, \cite{butter_publication}.
BUTTER includes performance metrics such as training and test loss for every epoch (3,000 epochs in most cases), and is partitioned into ``sweeps'' of the hyperparameter space, with a central ``primary sweep'' intersecting most of the other sweeps.
BUTTER-E adds both CPU and GPU energy measurements to the primary sweep, and due to the identical computational workflows, BUTTER-E measurements can be applied to the ``label noise", ``learning rate", and ``regularization" sweeps of BUTTER as well.
Because BUTTER-E sweeps the same experimental space as the primary sweep of the BUTTER dataset, we refer the reader to the BUTTER publication, \cite{butter_publication}, and dataset, \cite{butter_oedi_osti_dataset}, for additional details concerning exact hyperparameters tested.
Both BUTTER-E and BUTTER are freely available via the Open Energy Data Initiative (OEDI) Data Lake. 

The energy measurements for BUTTER-E were created by re-running the primary sweep of the BUTTER dataset on NREL's Eagle HPC system using a dedicated node for each training run.
Eagle's system configuration is detailed in Table \ref{tab:hpc_nodes} and the specifications of the CPUs and GPUs used are listed in Table \ref{tab:spec_comparison}.
The instantaneous power consumption of each training run was measured by the integrated Hewlett-Packard Enterprise Integrated Lights-Out chip at 1-minute intervals.
CPU runs were executed on Eagle's CPU-only nodes using Intel's OneDNN and Tensorflow packages using Intel's recommended settings. 
GPU runs were executed on Eagle's dual GPU nodes using NVIDIA's recommended settings and Tensorflow's parameter mirroring strategy.
One Tensorflow CPU thread was used for each of the two GPUs and each thread was pinned to a different core of a single CPU socket.
The raw, 1-minute resolution power data was then joined temporally with the scheduler information pertaining to the BUTTER-E runs as described in Section \ref{sec:data_quality}. The resultant run-level energy consumption data, as well as the 1-minute time series of instantaneous power measurements, and node-level power quantile estimates, are all published as part of the BUTTER-E dataset.

\begin{table}[t]
    \caption{\textbf{Specifications of the HPC nodes used to generate BUTTER-E.}}
    \label{tab:hpc_nodes}
    \centering
    {
    \setlength\tabcolsep{1.0mm}
    \begin{tabular}{|c|rcrcr|r|}
        \midrule
         Name & \# & CPU & RAM & GPU & SSD & Idle Power\\
         \midrule
         cpu1 & 1800 & 2x Xeon 6154 & $\SI{96}{\gibi\byte}$ & - & $\SI{1}{\tera\byte}$& $\SI{220}{W}$\\
         cpu2 & 720 & 2x Xeon 6154 & $\SI{192}{\gibi\byte}$ & - & $\SI{1}{\tera\byte}$& $\SI{230}{W}$\\
         cpu3 & 38 & 2x Xeon 6154 & $\SI{768}{\gibi\byte}$ & - & $\SI{1.6}{\tera\byte}$& $\SI{309}{W}$\\
         cpu4 & 10 & 2x Xeon 6154 & $\SI{768}{\gibi\byte}$ & - & $\SI{25.6}{\tera\byte}$& $\SI{388}{W}$\\
         gpu1 & 40 & 2x Xeon 6154 & $\SI{768}{\gibi\byte}$ & 2x V100 & $\SI{1.6}{\tera\byte}$& $\SI{374}{W}$\\
         gpu2 & 10 & 2x Xeon 6154 & $\SI{768}{\gibi\byte}$ & 2x V100 & $\SI{25.6}{\tera\byte}$& $\SI{403}{W}$\\
         \midrule
    \end{tabular}
    }
\end{table}
\begin{table}[t]
    \caption{\textbf{CPU and GPU specifications.}}
    \label{tab:spec_comparison}
    \centering
    {
        \renewcommand\footnoterule{}     
        \begin{tabular}{|c|c|c|}
            \hline
            \textbf{Property} & \textbf{Xeon Gold 6154} & \textbf{V100 PCIe} \\
            \hline
            \DeclareSIUnit\core{core}
            \DeclareSIUnit\sm{\textsc{SM}}
            Manufacturer & Intel & NVIDIA \\
            TDP & $\SI{200}{W}$ & $\SI{250}{W}$ \\
            Base Frequency& $\SI{3.0}{GHz}$ & $\SI{1.2}{GHz}$ \\
            Boost Frequency & $\SI{3.7}{GHz}$ & $\SI{1.53}{GHz}$ \\
            Memory Bandwidth & $\SI{128}{\gibi\byte/s}$ & $\SI{897}{\gibi\byte/s}$ \\
            Memory/Package & $\SI{48}{\gibi\byte}$-$\SI{384}{\gibi\byte}$ & $\SI{16}{\gibi\byte}$\\
            Cores/SMs & 18 & 84 \\
            L1 Data Cache & $\SI{32}{\kibi\byte/core}$ & $\leq$ $\SI{96}{\kibi\byte/SM}$\\
            L2 Cache & $\SI{1}{\mebi\byte/core}$ & $\SI{6}{\mebi\byte}$ shared \\
            L3 Cache & $\SI{24.75}{\mebi\byte}$ shared & - \\
            FP32\footnote{\scriptsize We abbreviate single-precision 32-bit floating-point numbers as ``FP32" here.} Vector Width & 16 (AVX-512) & 64 FP32 cores/SM\\
            Register File Size & $\SI{2}{\kibi\byte/core}$ \footnote{\scriptsize 32 AVX-512 FP32-compatible registers per thread.
            Simultaneous multi-threading (SMT) was disabled in these experiments. } & $\SI{256}{\kibi\byte/SM}$\\ 
            \hline
        \end{tabular}
        \vspace{-2ex}
    }
\end{table}

\section{Data Quality and Processing}
\label{sec:data_quality}

\begin{figure}[h!]
    \centering
    \includegraphics[width=\linewidth]{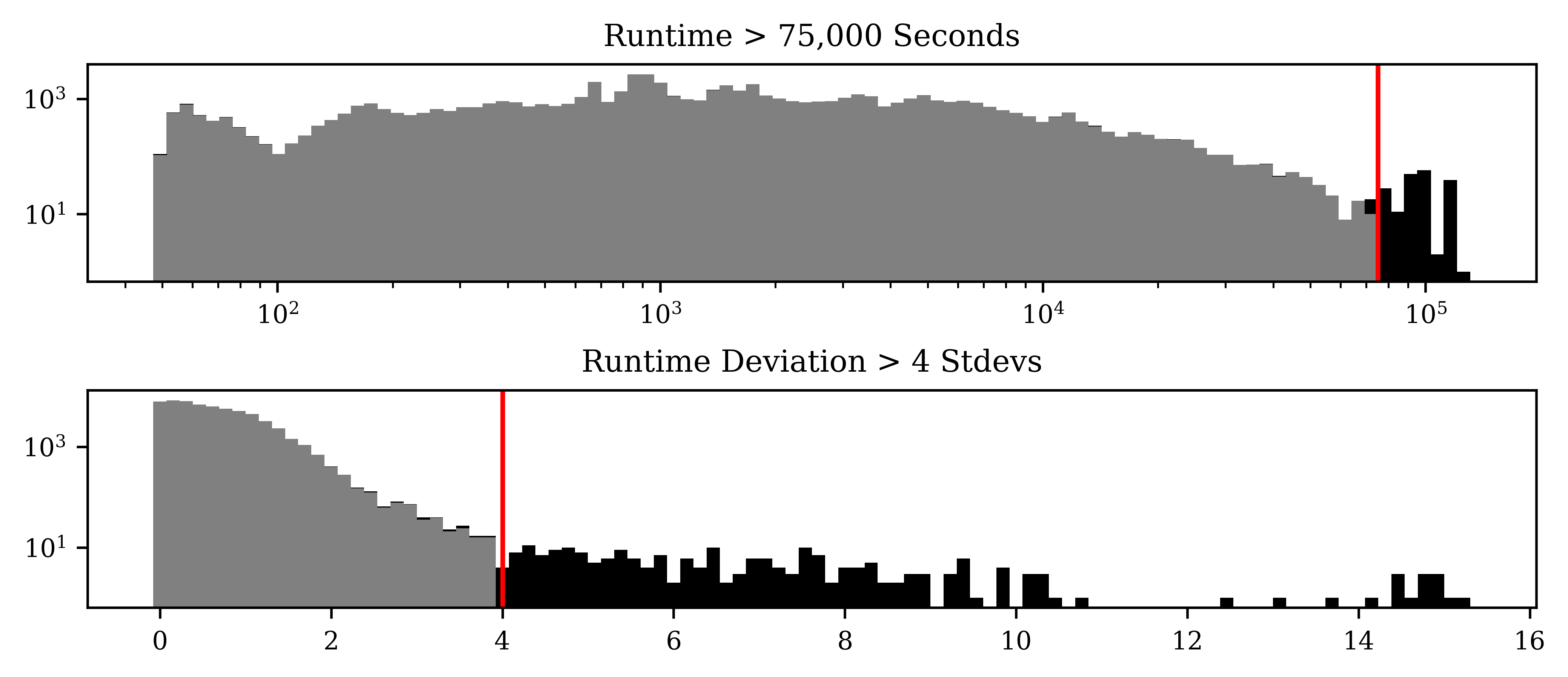}
    \caption{Histograms (on logarithmic axes) showing the quantity and location of data filtered out for this analysis. The filters reject 241 runs, which is approximately 0.6\% of the total number of runs.
    }
    \label{fig:filtering}
\end{figure}



The raw, minute-level time-series power data from the node-level watt-meters was joined with run data from the HPC scheduler, matching the job ID to the node name and time for which the job was run. 
For each run, the power data was integrated over the length of time for which a given run was utilizing each of its compute nodes using a linear interpolation to deal with endpoints. 
For the interior data points (the vast majority of data points) this degenerates into a simple sum and unit conversion.

We apply some basic data quality filters during this stage in the data pipeline.
In total, 665 out of 41,129 runs (1.6\% of all runs) were excluded here due to either 0W readings present in the data (470 runs) or missing data (195 runs).
Of the 40,464 training runs that did join successfully with energy data, we continue to perform more filtering to remove low-quality and outlier data points.
Initial visual inspection revealed a cluster of runs with very long, unrealistic, run-times greater than 75,000 seconds. We filter these out as shown in Figure \ref{fig:filtering}a.
There were 163 runs matching this criterion, and these were removed from the data.
Further inspection revealed that most of these long runs were started on a specific day, which further points to a hiccup in the underlying system rather than a real effect. Our analyses indicate that the system was likely experiencing technical difficulties during this time.

We also remove statistical outliers in runtime with respect to training runs with similar hyperparameter combinations. 
The goal is to filter out runs for which energy measurements were unsuccessful, or for which the HPC was running in an abnormal state and where the runs may have taken longer than can be reasonably expected. 
This filter rejects any run with a log runtime greater than 4 standard deviations from the mean log runtime for runs with the same dataset, network size, and whether it ran on a GPU or CPU node.
This filter discards an additional 78 runs. 
The histogram of the runtime standard deviations filter is presented in Figure \ref{fig:filtering}b, with the filter criteria indicated. Combined, the two filters reject 241 runs (0.6\% of the total number of runs with energy data), resulting in 40,223 usable runs.

One challenge that arises when comparing energy measurements taken on different computing nodes are the differences in each system's idle power draw.
For example, when idling (but not in a low-power sleep state) a cpu4 node consumes $\SI{168}{W}$ more power than a cpu1 node that is identical except for the quantity of RAM and type of SSD installed. (See Table \ref{tab:hpc_nodes} for the definitions of these node types).
These differences are effectively integrated over the duration of an experiment meaning that if identical workloads are executed on cpu1 and cpu4 nodes over an hours time, the cpu4 node will measure $\SI{605}{kJ}$ more energy consumed.
To control for these biases and allow us to compare apples-to-apples across runs executed on different node types, we estimated the energy that would be consumed on a cpu1 node for CPU experiments, and a gpu1 node for GPU experiments.
To achieve this standardization, we first estimated each node type's idle power draw by using the $2\%$ quantile of the node type's power draw distribution as measured over a nine-month timeframe.
Next, we subtracted the differences between the idle power draw of the node each experiment ran on multiplied by experiment runtime from the measured energy to arrive at the ``standardized energy'' of each experiment.

A second challenge arose when comparing energy per batch and per datum across different sized datasets.
Experiments on smaller datasets tended to take less time, making experimental setup and tear-down a larger component of total energy consumption.
To remove this confounding interaction from our analysis, we estimated the experimental energy and runtime overheads as being the $5\%$ quantile of the runtime and standardized energy over all experiments using the very small tecator\_505 dataset and with $2^{10}$ or fewer trainable parameters.
The estimated standardized overheads were $\SI{14.0}{kJ}$ and $\SI{52.9}{s}$ for CPU experiments and $\SI{27.2}{kJ}$ and $\SI{656}{s}$ for GPU experiments.
For small models and small datasets, these overheads are non-negligible and can be important to consider.
We subtracted the overheads for our analysis of aggregate trends in Section \ref{sec:aggregate_trends} and cache effects \ref{sec:cache_effects}, but use the full standardized energy when fitting the energy model in Section \ref{sec:energy_model}.


\section{Aggregate Trends}
\label{sec:aggregate_trends}
\begin{figure*}[tb]
    \centering
    \includegraphics[width=\textwidth]{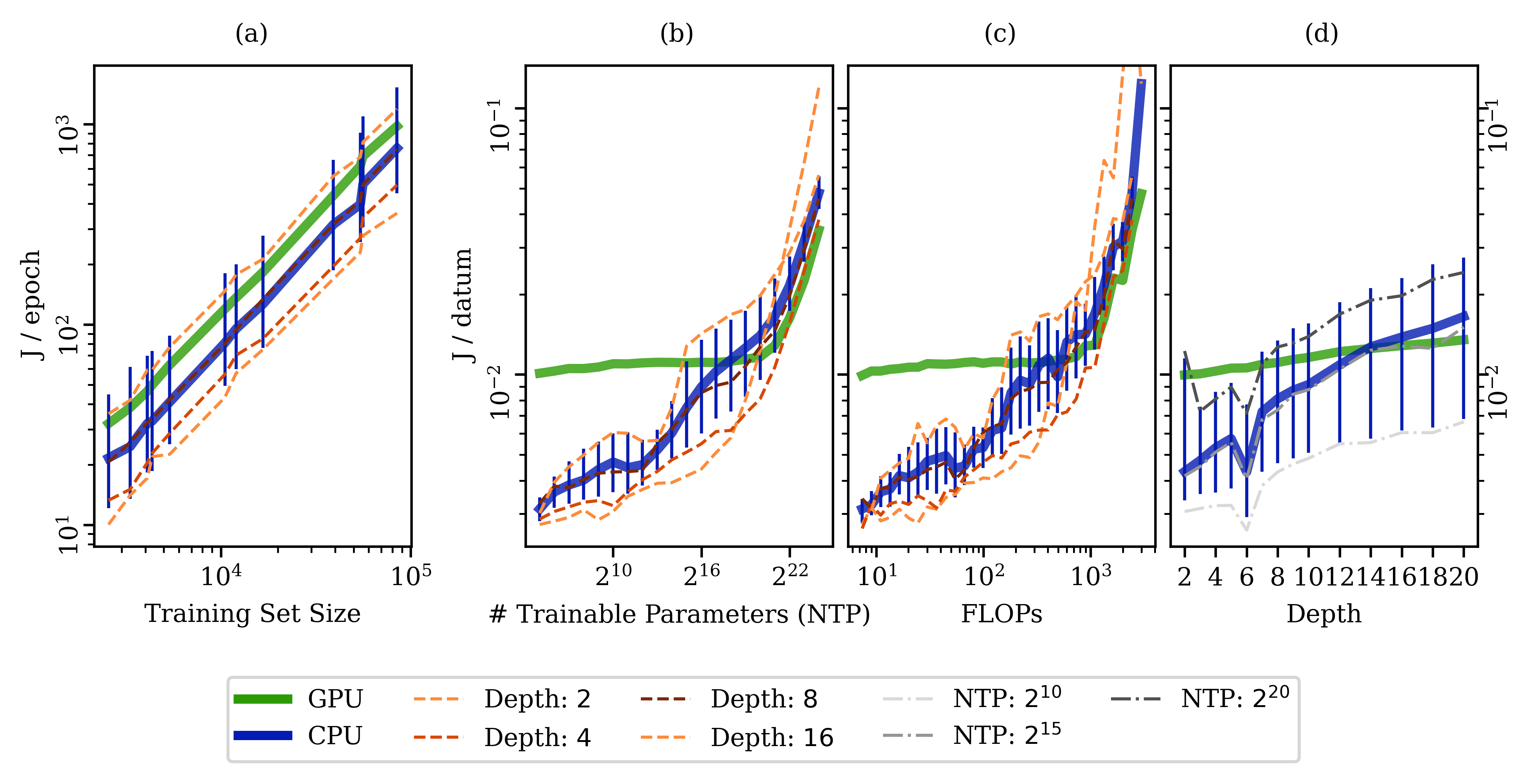}
    \caption{(a) Illustrates the linear relationship between training set size and energy consumed per-epoch. 
    (b) Shows the marginal effect of number of trainable parameters (NTP) on energy per training datum per epoch.
    NTP has a positive nonlinear relationship to energy used.
    (c) Shows the same for the effect of FLOPs.
    FLOPs has a positive nonlinear relationship to energy used that is very similar to NTP's relationship to energy.
    (d) Shows the same for the effect of Depth. 
    Depth and energy per datum are also is positively related; we observe that deeper networks use more energy per batch.
    }
    \label{fig:aggregate_trends}
\end{figure*}
In this section, we discuss the energy consumption patterns marginalized across the BUTTER-E dataset, using the median and interquartile ranges to aggregate the data.
In Figure \ref{fig:aggregate_trends}a, we observe that the energy per training epoch (an epoch is a complete pass through all of the training data) is linearly related to the number of data points processed in each epoch.
Therefore, to focus on the impact of other parameters on energy consumption, we examine joules per training datum in this section (that is total data points processed, including repetitions of the same datum when it is reused each epoch), rather than joules per epoch.
Note that the number of batches is also, aside from rounding up, proportional to the amount of training data.
In this case, GPU-based training consumed a median marginal cost of $9.47 \si{mJ/datum}$ with an upper quartile (UQ) of $13.75 \si{mJ/datum}$ and a lower quartile (LQ) of $7.7 \si{mJ/datum}$. 
While CPU based training consumed $6.16 \si{mJ/datum}$ with an  UQ of $16.32 \si{mJ/datum}$ and LQ or $2.41 \si{mJ/datum}$. 
This may be somewhat surprising, as it shows that our GPUs were actually less energy efficient than CPUs in this respect.

Now, let's investigate the marginal effect of the number of trainable parameters (NTP), floating point operations (FLOPs), and the network depth on energy per training datum.
One might expect that energy per datum would be linearly related to the number of FLOPs per datum, or to NTP.
This could be a reasonable assumption, as FLOPs are a fundamental unit of computation, and the number of trainable parameters in a network is closely related to this quantity.
However Figures \ref{fig:aggregate_trends}b and c show that neither relationship appears to be linear in our dataset.
In GPU experiments, the energy per datum did not increase substantially until NTP reaches $~2^{20} \approx 1M$ where it then enters a fairly linear regime.
$2^{20}$ parameters coincides closely with the \SI{6}{\mebi\byte} capacity of the GPUs L2 cache, enough to contain $1.5 \times 2^{20}$ 32-bit floating point parameters, exactly half-way between the $2^{20}$ and $2^{21}$ parameter counts tested in the dataset.
CPU experiments see an overall increase in energy per parameter, but the rate of increase begins slowly and increases in a non-linear fashion, ultimately settling into a linear relationship near the same $2^{20}$ parameter count.
This non-linear response is likely due to several factors including caching effects, which we explore in the next section, a per-epoch overhead, and for very small networks, potentially falling short of the vectorization or parallelization width of the hardware.
In the linear regimes, we see a median marginal cost of $6.038 \si{mJ/datum}$ for CPU runs and $9.39 \si{mJ/datum}$ for GPU runs.
In CPU experiments 20-layer networks consumed a median of $2.101 \si{nJ/datum}$, and $1.495 \si{nJ/datum}$ on GPUs.
\textbf{In our dataset, neither number of trainable parameters nor number of FLOPs were shown to have a linear effect on energy efficiency.}

Depth is a salient factor in determining the energy costs of neural networks and the optimal hardware architecture on which to train them.
Figure \ref{fig:aggregate_trends}d highlights a clear positive relationship between depth and energy per training datum across a wide range of parameter counts.
20-layer CPUs consumed a median of $9.94 \si{mJ/datum}$, while 2-layer CPU networks has a median of $3.02 \si{mJ/datum}$.
While not an entirely linear relationship, adding a layer to a network had a median marginal cost of $3.865 \si{mJ/datum}$ on CPUs and a considerably lower $2.03 \si{mJ/datum}$ on GPUs.
In these experiments, GPUs were less energy efficient than CPUs on shallow networks but more efficient on deep networks.
This may be due in part to higher per-layer loop and/or call overheads, such as calls to the underlying cuDNN or oneDNN libraries, for CPUs.
It may also be due to different cache access patterns, such as those used to pass the outputs of one layer into the inputs of another.
We explore the per-layer energy costs on CPUs and GPUs further in Section \ref{sec:energy_model}.

\section{Cache Effects}
\label{sec:cache_effects}
\begin{table}[tb]
    \caption{\textbf{Symbols Used}}
    \label{tab:symbols}
    \centering
    {
    \setlength\tabcolsep{1.0mm}
    \begin{tabular}{|c|c|p{0.6\columnwidth}|}
        \midrule
         Symbol & Units & Description \\
         \midrule
         $n$ & epoch & \# training epochs \\
         $h_t$ & batch/epoch & \# training batches per epoch \\
         $h_s$ & batch/epoch & \# test batches per epoch \\
         $L$ & - & Set of network layers \\
         $|l|$ & units & Number of units in layer $l \in L$\\
         $a_l$ & - & $\mathbb{R}^{|l|\times 1}$ activation vector for layer $l$\\
         $z_l$ & - & $\mathbb{R}^{|l|\times 1}$ pre-activation vector for layer $l$\\
         $\mathrm{g}_l(z_l) = a_l$ & - & Layer $l$'s activation function\\
         $\Theta_l$ & - & $\mathbb{R}^{(|l-1| + 1)\times |l|}$ matrix of network parameter values for layer $l$ \\
         $\Theta$ & - & $\{\Theta_l \forall l \in L\}$ Set of all network parameters\\
         $\mathcal{L}$ & - & Loss \\
         $C$ & - & Set of memory levels in ascending order, \{L1, L2, L3, RAM\}\\
         $c_x \in C$ & - & Memory level of working set $x$\\
         $s_x$ & bytes & Size of working set $x$ \\
         $d$ & - & Batch input/output working set\\
         $f$ & - & Forward pass working set during training\\
         $f'$ & - & Forward pass working set during testing\\
         $b$ & - & Backward pass working set\\
         $T$ & - & $\{t_l \forall l \in L\}$ inter-layer working sets\\
         $E_t$ & J & Total energy cost of an experiment\\
         $E_x$ & J & Total energy cost of a pass $x \in \{f,b,f'\}$\\
         $o_x$ & FLOPS & Operations executed in pass $x \in \{f,b,f'\}$\\
         $k_e$ & J & Energy overhead of an experimental run\\
         $k_p$ & J & Energy overhead of a pass\\
         $k_o$ & J/FLOP & Marginal energy per operation \\
         $k_d$ & J/layer & Marginal energy per layer \\
         $a_x$ & J & Transfer initiation energy to access memory level $x$ \\
         $m_x$ & J/byte & Marginal energy per byte accessed from memory level $x$ \\
         $\phi(x)$ & J & Energy required to access working set $x$\\
         \midrule
    \end{tabular}
    }
\end{table} 
\begin{figure}[h!]
    \centering
    \includegraphics[width=\linewidth]{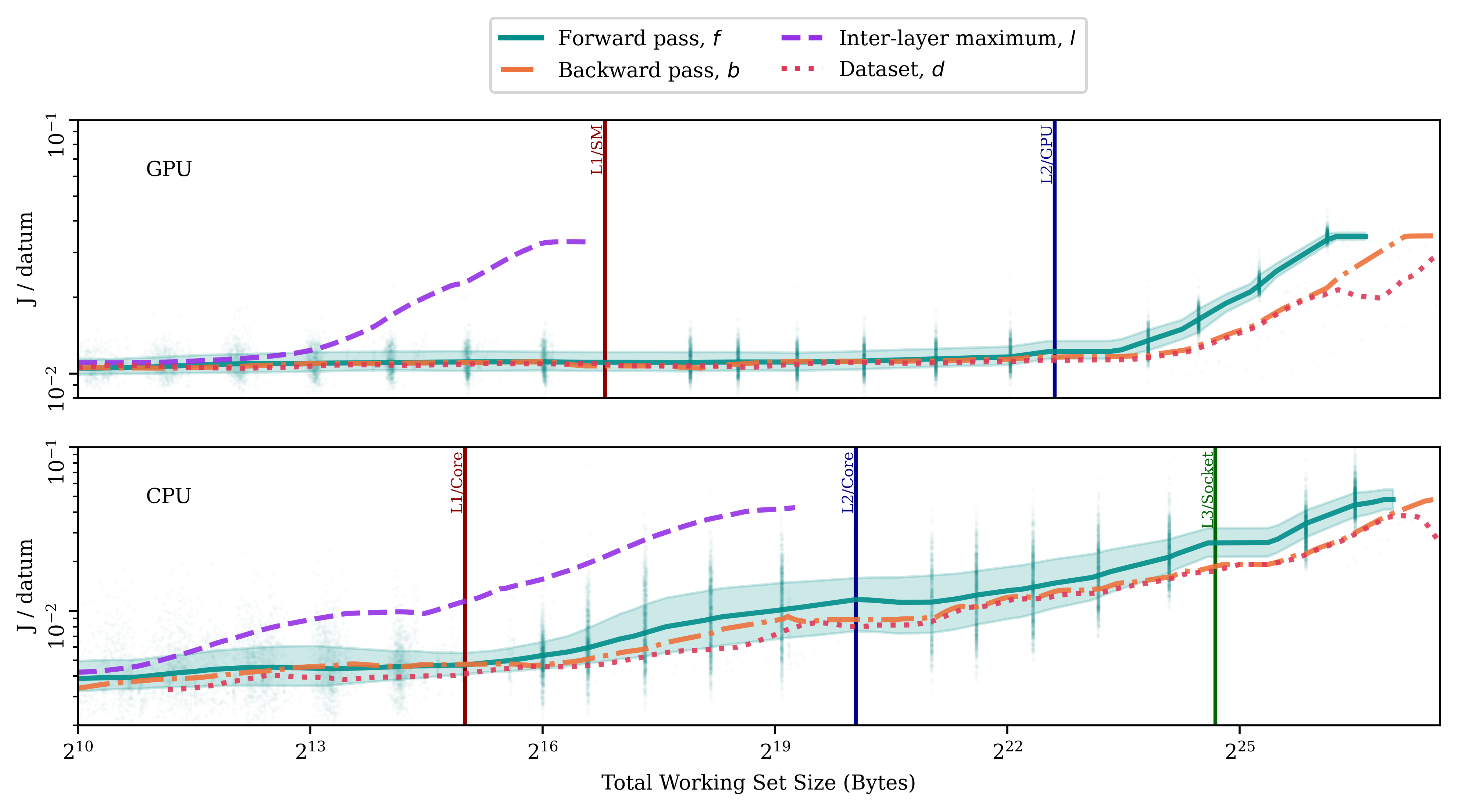}
    \caption{(a) GPU and (b) CPU energy consumption as a function of working set size for four key working sets. The vertical lines spanning each subplot indicate the size of each physical cache level in the hardware. Energy consumption increases appear to coincide with certain working sets spilling into higher memory levels.
    We aggregate the energy consumption by using a windowed median for each working set, as a function of effective working set size, using a window size of $\pm 50\%$.
    For the Forward Pass only, the grey filled area spans the interquartile range and alpha-blended points show individual measurements.
    }
    \label{fig:cache_effects}
\end{figure}
In this section we dive deeper into the computational workflow being executed during training to better understand the non-linear effect of number of trainable parameters and FLOPs, and to uncover the factors that impact the energy used to train the networks in our dataset.
The experimental workflow executed during each experiment is described listed in Algorithm \ref{alg:experimental_procedure}.
To this end, we introduce several \emph{working sets}, which we define as sets of intermediate data used in the computation which is stored in either the cache or system memory. 
The size of each working set varies based on the architecture of the network.
For example, networks with wider layers require more memory to hold the weights, biases, activations, and gradients relevant to the computation between each layer.
This notion is captured in our \emph{inter-layer working set}, which we define below among the others.
We compute key working sets for each experiment and compare them to the hardware cache sizes, providing a reasonable estimate of how well a given experiment is able to fit into each cache level. 
While the exact working sets critical to network energy usage will vary depending on implementation details, we define four key working sets, listed below, in this analysis.
Table \ref{tab:symbols} provides an index of the symbols used in this section and beyond. The four key working sets are:
\begin{itemize}
    \item The {\bf Forward-pass working set}, $f$ contains only the model parameters, $\Theta$.
    We use $f$ to describe the forward pass working set during inference and $f'$ to describe it during training.
    While $f$ and $f'$ are identically sized, we distinguish them because $f'$ may be pushed to a higher memory level because $f'$ usage alternates with with $b$ during training.
    \item The {\bf Backward-pass working set}, $b$ contains parameters $\Theta$, their gradients $\frac{\partial \mathcal{L}}{\partial \Theta}$, unit pre-activations $z_l \forall L$, and unit gradients $\frac{\partial \mathcal{L}}{\partial a_l} \forall l \in L$.
    Because unit pre-activations can be supplanted by gradients as backpropagation proceeds, we approximate the size of the backward pass working set as only containing one of the two values per unit at any time.
    Note that there is a distinct unit activation and gradient for every datum of the batch.
    \item The {\bf Inter-layer working set}, $t_l \forall l \in L$ contains the input and output activations or activation gradients to each layer. 
    Equal in size to the number of units in the layer plus the number of units in the previous layer multiplied by the batch size.
    During forward passes the inter-layer working set consists of unit activations $a_{l-1}$ and $a_l$, in backward passes it consists of activation gradients $\frac{\partial \mathcal{L}}{\partial a_{l}}$ and $\frac{\partial \mathcal{L}}{\partial a_{l-1}}$.
    \item The {\bf Dataset working set}, $d$ contains the training and test datasets only.
\end{itemize}

In Figure \ref{fig:cache_effects}, we explore the energy consumed per training datum as a function of each of the four key working set sizes.
Here, we observe that marginal energy consumption increases when $f$ and $b$ spill out of a cache level.
Although not visible in the figure, and very slight, this is even true when $f$ and $b$ spill from L1 into L2 cache on the GPU.
In Section \ref{sec:energy_model}, we fit an energy model to the experimental data that suggests that the number and size of transfers at each memory level are the critical factors.
In this case both $f$ and $b$ determine energy consumption with $b$ causing somewhat greater energy usage than $f$ due to its larger size.
However, because the effects of both $f$ and $b$ are observed concurrently at a limited number of NTPs, additional experimentation would be required to definitively ascertain the energy costs of each working set.
While the impact of the per-layer working set, $t_l$, spilling from the cache is less distinct than the impact of the forward or backward working sets (maximum size is plotted in Figure \ref{fig:cache_effects}), energy consumption rose substantially just past the points at which the largest per-layer forward and backward working sets spill into L2.
And finally the dataset working set, $d$, does not have an immediately distict impact in this figure, but once $f$, $b$, and $l$ were accounted for in our energy model regression as described in Section \ref{sec:energy_model}, including $d$ markedly reduced regression error in cases where where $d$ resided in a higher memory level than $f$ and $b$.
In these cases, per-batch accesses to $d$ are responsible for the majority of accesses to the highest memory level, causing it to play an important role in energy consumption.

\section{An Energy Model for MLPs}
\label{sec:energy_model}
In this section, we combine the insights from Sections \ref{sec:aggregate_trends} and \ref{sec:cache_effects} to build a model of energy consumption based on the hyper-parameter choices, taking into account the non-linear interactions of the working sets and hardware cache sizes. 
Practitioners might assume that energy consumption per training batch increases linearly with the number of operations per training batch, and by connection the number of trainable parameters (NTP) in the network.
In contrast to this common-sense assumption, and as discussed in Section \ref{sec:aggregate_trends}, we observed that while energy per training datum increases with FLOPs and NTP, the relationship is actually nonlinear, mediated in part by the interactions between cache size and working set sizes.
Once working set sizes and depth were taken into account, we observed no clear additional impact of other hyperparameters, such as dataset, network shape, NTP.

\begin{algorithm}
\caption{
A sketch of the computational workflow executed during training. 
Regions corresponding to the four key working sets are parenthetically annotated.
}
\label{alg:experimental_procedure}
\begin{algorithmic}[1]
\For{each training epoch $n$}
    \For{the $h_t$ batches in the training dataset}
        \State Load training data for batch $(d)$
        \State \textbf{Training Forward Pass $(f)$:}
        \For{each layer $l \in L$ from input to output $(t_l)$:}
            \State Compute activations:
            \State \(z_{l} = \Theta_{l} \cdot [a_{l-1}, 1] \)
            \State \(a_{l} = \mathrm{g}(z_{l})\)
        \EndFor
        \State Compute and accumulate batch training loss \(\mathcal{L}\)
        \State Compute output gradient \(\frac{\partial \mathcal{L}}{\partial a_{l}}\)
        \State \textbf{Backward Pass $(b)$:}
        \For{each layer $l \in L$ from output to input $(t_l)$:}
            \State Compute parameter gradients:
            \State \( \frac{\partial \mathcal{L}}{\partial \Theta_{l}} = \frac{\partial \mathcal{L}}{\partial a_{l}} \cdot [a_{l-1}, 1]^T\)
            \State Compute input activation gradients:
            \State \(\frac{\partial \mathcal{L}}{\partial a_{l-1}} = (\Theta_{l}^T \cdot \frac{\partial \mathcal{L}}{\partial a_{l}}) \cdot \mathrm{g}'(z_{l-1})\)
        \EndFor
        \State Update \(\Theta\) using \(\frac{\partial \mathcal{L}}{\partial \Theta}\)
    \EndFor
    \State \textbf{Testing Forward Pass $(f')$:}
    \For{the $h_s$ batches in the test dataset}
        \State Load feature and response data for this batch $(d)$
        \For{each layer $l \in L$ from input to output $(t_l)$:}
            \State Compute activations:
            \State \(z_{l} = \Theta_{l} \cdot [a_{l-1}, 1] \)
            \State \(a_{l} = \mathrm{g}(z_{l})\)
        \EndFor
        \State Calculate and accumulate batch test loss \(\mathcal{L}\)
    \EndFor
\EndFor
\end{algorithmic}
\end{algorithm}
We start by ordering the working sets based on how frequently they are accessed, and sorting them into the lowest cache level they will fit into. 
The inter-layer working sets, $T = \{t_l \forall l \in L\}$, contain the data passed from layer to layer and are accessed in the inner most loops of Algorithm \ref{alg:experimental_procedure} and is the same size in forward and backward passes.
In this energy model we place the inter-layer working set into the lowest memory level it would fit into.
We approximate the inter-layer working set as being evenly distributed across all processing units (as might be accomplished by parallelizing the computation for each member of the batch).
For example, our compute nodes each have 36 total CPU cores, each with separate L1 and L2 caches, so we model the inter-layer working set as residing in L1 cache when $s_l / \SI{36}{cores} \leq \SI{64}{\kibi\byte}$.

Per-layer portions of $f$, $f'$, and $b$, $\Theta_l$ and $\frac{\partial \mathcal{L}}{\partial \Theta_l}$, are accessed on each inner loop iteration.
However, because these per-layer portions are accessed sequentially, we approximate $f$, $f'$, and $b$ as residing in the lowest memory level that can accommodate both the largest inter-layer working set $t_l$ and the entirety of the batch working set ($f$, $f'$, or $b$).
If some caches are exclusive, this approximation can be adjusted to account for the larger effective cache capacity available.
We approximate $f$ and $f'$ as being replicated among all of the cores or SMs in the system.
For example, if the entirety of $\Theta_l$ does not fit in a single core's L1 cache, we push $f$ into L2 cache, and so on.
As with $f$ and $f'$, we approximate the parameters and parameter gradients of $b$ as being replicated, but model the unit pre-activations and gradients in $b$ as being evenly distributed among the cores or SMs of the system.

Finally, the dataset working set, $d$, is accessed once per batch to load the next batch of data.
Because $d$ is accessed in least-recently-used order and cycled through once per epoch we consider accesses to the dataset working set to reside in the lowest memory level that can accommodate the inter-layer, backward pass, and dataset working sets.
As with the inter-layer working set, we approximate $d$ as being evenly distributed across all processing units in the system.
Based on this reasoning, we can rank the size of each working set as follows:
\begin{equation}
    c_l \leq c_{f'} \leq c_f = c_b \leq c_d
\end{equation}
Using these four working sets and our method for approximating the memory level in which each set resides, we adopt an affine model for estimating the energy cost of accessing a working set $x$ (reading or writing) from level $c$ of the memory hierarchy.
\begin{equation}
    \phi(x) =  a_c + m_c s_x\quad\text{s.t.}\quad c_x = c \in C
\end{equation}
With this, we then model the energy cost of a network pass (forward or backward) as the sum of the data transfer costs of accessing each working set $\phi$, a per-pass overhead $k_p$, and a per-operation cost $k_o$, and a per-layer overhead $k_d$
\begin{align}
    E_f     &= k_p + k_o o_f + k_d |L| + \phi(d) \nonumber\\
            &\quad + \phi(f) + \sum_{t_l \in T} \phi(t_l)\\
            &= k_p + k_o o_f + k_d |L| + \sum_{x \in \{d,f\} \cup T} \phi(x)\\
    E_{f'}  &= k_p + k_o o_f + k_d |L| + \sum_{x \in \{d,f'\} \cup T} \phi(x)\\
    E_{b}   &= k_p + k_o o_b + k_d |L| + \sum_{x \in \{b\} \cup T} \phi(x)
\end{align}
Combining these per-pass energy costs into a per experiment energy cost, we arrive at:
\begin{align}
\hat{E_t} &= k_e + n \left( h_t \left( E_f + E_b \right) + h_s E_{f'} \right) \\
&= k_e \nonumber\\
&\quad + n (2h_t+h_s) k_p \nonumber\\
&\quad + n\left(h_t+h_s\right) o_f k_o \nonumber\\
&\quad + n h_t o_b k_o \nonumber\\
&\quad + n (2h_t+h_s) |L| k_d\nonumber\\
&\quad + n (h_t+h_s) \phi(d) \nonumber\\
&\quad + n h_t \phi(f) \nonumber\\
&\quad + n h_t \phi(b) \nonumber\\
&\quad + n h_s \phi(f') \nonumber\\
&\quad + n (2h_t+h_s)\sum_{t_l \in T} \phi(t_l)
\end{align}
where $n$ is the number of training epochs, $t$ the number of training batches per epoch, and $s$ the number of test batches per epoch, and $k_e$ denotes an additional per-run energy overhead (one constant for each type of hardware, CPU or GPU).
Because $\phi(\cdot)$ is an affine function, we can consolidate terms to produce a concise expression for the total modeled energy:
\begin{equation}
    E_t = \textbf{w}^T Q \textbf{k}
\end{equation}
where $\textbf{w} = [1, n (2h_t+h_s), n\left(h_t+h_s\right), \dots] \in \mathbb{R}^{(9+|L|)\times 1}$ is a vector of cost coefficients, $Q \in \mathbb{R}^{(9+|L|)\times(4+2|C|)}$ is an experiment-dependant matrix weighting and mapping each coefficient onto the cost vector, $\textbf{k} = [k_e, k_p, k_o, k_d, a_1 \dots, m_1 \dots] \in \mathbb{R}^{(4+2|C|)\times 1}$.
Using this form, we computed the estimated size and memory levels of each working set for every experiment in the BUTTER-E dataset and fit a least squares regression with parameters $\textbf{k}$ for both CPU and GPU experiments.
As measured $E_t$ ranged from $30\si{kJ}$ to $20\si{MJ}$, minimizing squared or absolute error would cause the model to be biased in favor of high-energy experiments.
Instead, we fit $\textbf{k}$ to minimize $\sum(\log(\hat{E_t} / E_t))^2$ which aims to identify a model for which $<\hat{E_t} / E_t> \approx 1$.
\begin{table}[tb]
\centering
\caption{Fitted model over all CPU experiments and GPU all experiments.}
\label{table:regression_parameters}
\begin{tabular}{|crrll|}
\hline
Parameter & CPU & GPU & Units & Description\\
\hline
$k_e$ & $10200$ & $272000$ & \si{J/experiment} & Per-experiment energy overhead\\
$k_p$ & $1460$ & $3080$ & \si{J/pass} & Per-pass loop energy overhead\\
$k_o$ & $744$ & $890$ & \si{J/GFLOP} & Energy cost of one billion operations\\
$k_d$ & $\approx 0$ & $33.0$ & \si{J/layer} & Per-layer call/setup overhead  \\
$a_1$ & $\approx 0$ & $16.4$ & \si{J/access} & L1 access energy overhead\\
$a_2$ & $59.3$ & $19.8$ & \si{J/access} & L2 access energy overhead\\
$a_3$ & $215$ & - & \si{J/access} & L3 access energy overhead\\
$a_4$ & $305$ & $\approx 0$ & \si{J/access} & RAM/VRAM access energy overhead\\
$m_1$ & $\approx 0$ & $\approx 0$ & \si{J/\mebi\byte} & Energy cost per byte accessed in L1 cache \\
$m_2$ & $23.0$ & $\approx 0$ & \si{J/\mebi\byte} & Energy cost per byte accessed in L2 cache\\
$m_3$ & $22.5$ & - & \si{J/\mebi\byte} & Energy cost per byte accessed in L3 cache\\
$m_4$ & $36.3$ & $5.7$ & \si{J/\mebi\byte} & Energy cost per byte accessed in RAM/VRAM\\
\hline
\end{tabular}
\end{table}

\begin{figure}[tb]
    \centering
    \includegraphics[width=\linewidth]{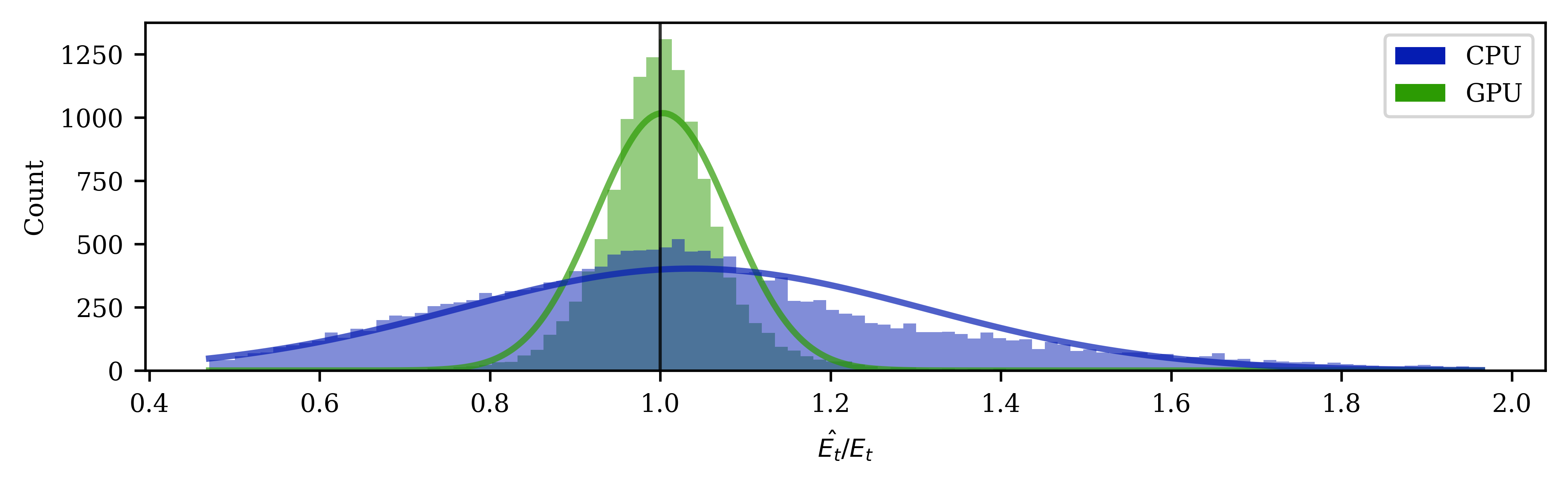}
    \caption{
    Error distribution of the ratio of model predicted to measured energy, $\hat{E_t} / E_t$.
    The estimates are unbiased regardless of the hardware type used, but the accuracy appears to be lower for experiments when trained on CPU.
    This makes some intuitive sense, since CPUs have more complicated cache hierarchies and instruction sets than GPUs.
    }
    \label{fig:error_distribution}
\end{figure}

The results of this regression are presented in Table \ref{table:regression_parameters}, and
the distributions of relative estimation error for both CPU and GPU hardware are shown in Figure \ref{fig:error_distribution}.
GPU experiment energy usage was predicted with an average error of $\pm 2.74\%$ and CPU energy usage with $\pm 20.2\%$.
Likely due to GPU node's higher baseline power consumption, GPU runs were better modeled as overheads $k_e$ and $k_p$ were considerably higher, resulting in lower measurement noise.
Surprisingly, the regression results suggest that CPU and GPU experiments have similar per operation energy costs despite quite different computational approaches.
In contrast to the aggregate trends of Figure \ref{fig:aggregate_trends}d, The regressed parameters indicate GPU experiments paid a non-negligible energy cost per network layer and per L1 access (primarily driven by accesses to the inter-layer working set and therefore model depth) while CPU experiments did not.
CPU experiments appeared to access L1 cache essentially for free but per-access and per-byte costs escalated up the memory hierarchy with a big step between L1 and L2 cache, a small step to L3, and a moderate additional energy cost to access system memory.
In comparison, GPU experiments showed somewhat small per-access costs for both L1 and L2 cache with no sensitively to access size.
In aggregate, while GPUs appeared to have higher per-layer energy overheads, lower per-access energy costs at L2 and higher memory levels resulted in the aggregate trends from Figure \ref{fig:aggregate_trends}d: CPUs consumed less energy than GPUs did in shallow networks, and more in deep networks.
However, due to CPUs lower per-layer overhead and L1 access costs, when layers were small enough that the inter-layer working set resided in L1 cache, CPUs were also more efficient than GPUs on deep networks.
In both CPU and GPU cases, working sets spilling into system memory incurred a substantial cost proportional to access size.

What conclusions can we draw from this?
For the CPU and GPU configurations tested, number of operations plays only a minor role in determining energy consumption of MLPs.
Instead, MLP energy consumption is closely tied to the interaction between working set sizes and the cache hierarchy.
As a rule-of-thumb, comparing parameter size and consecutive per-layer width to a system's cache size indicates what energy operating regime a MLP is likely to be in.
If the dataset does not fit into cache, energy consumption per batch will increase.
If the parameters or their gradients do not fit into cache, energy consumption increases linearly with the number of parameters.
If the inter-layer activations or activation gradients do not fit into cache, energy consumption increases linearly with layer width.
On the other hand, if a network fits into cache, energy consumption is almost constant without regard to network properties.
In the case of our GPU experiments, energy consumption was most closely tied to the number of parameters when the parameter working set spilled into GPU RAM.
CPU experiments revealed a richer memory hierarchy and therefore more complex effects with varying marginal costs at each memory level.
As we will see in the next section, these critical points where the pass and inter-layer working sets spill into higher memory levels  correspond to points of high energy-efficiency.


\section{The Energy-Loss Trade-off}
\label{sec:energy_efficiency}

\begin{figure}[p]
    \centering
    \includegraphics[width=\textwidth]{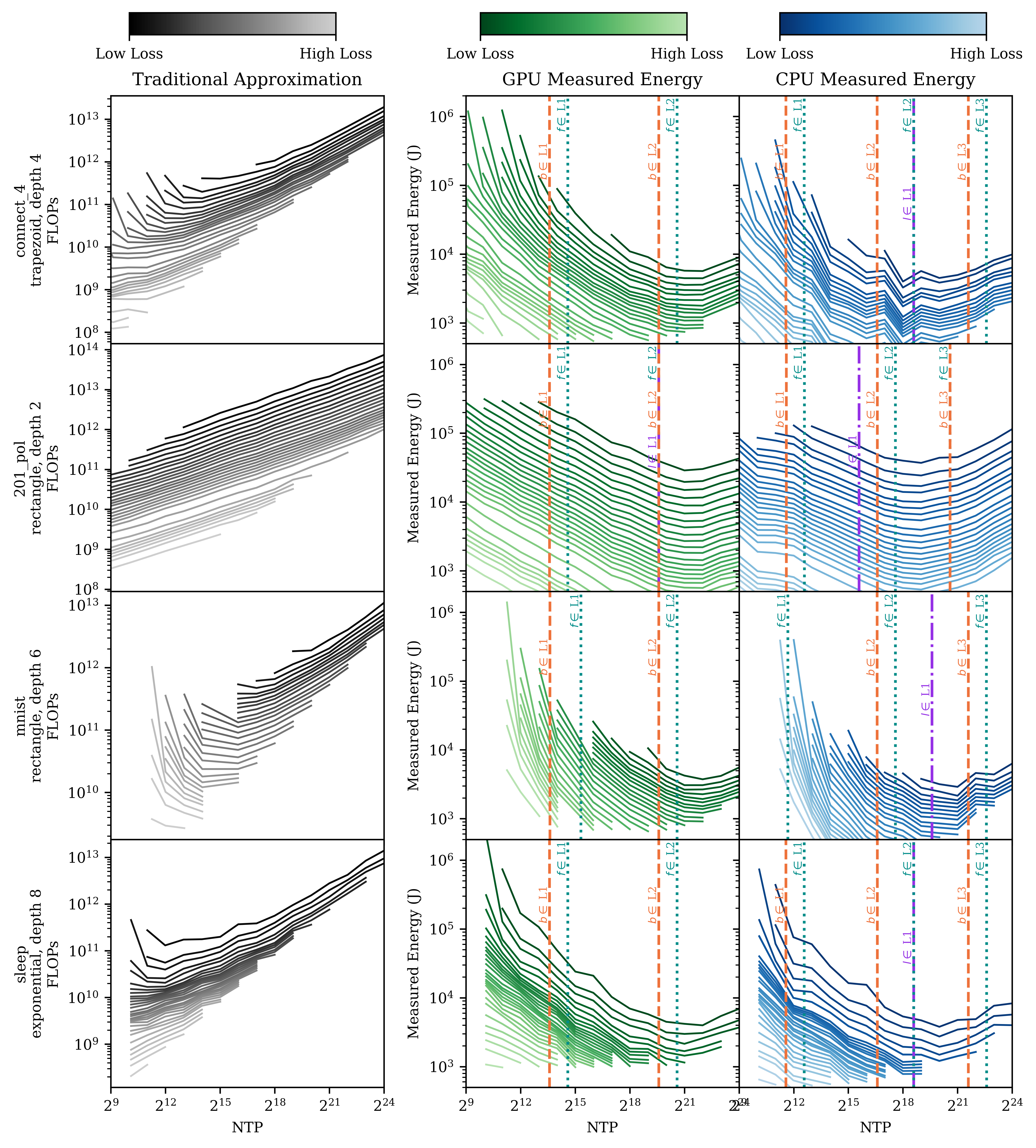}
    \caption{
    Comparing FLOPs as the traditional surrogate for relative energy consumption (left) against measured energy consumption on GPUs (middle) and CPUs (right) across a sample of datasets, network shapes, and depths.
    In the spirit of a topographic map, each isoloss line plots the median energy (or FLOPs) used to reach a test loss level.
    In contrast to operation count, we see that the most energy efficient NTP coincides with one or more working sets just spilling from a cache level.
    }
    \label{fig:sample_isolines}
\end{figure}

\begin{figure}[p]
    \captionsetup[subfigure]{labelformat=empty}
    \centering
    \begin{subfigure}
        \centering
        \includegraphics[width=.98\linewidth]{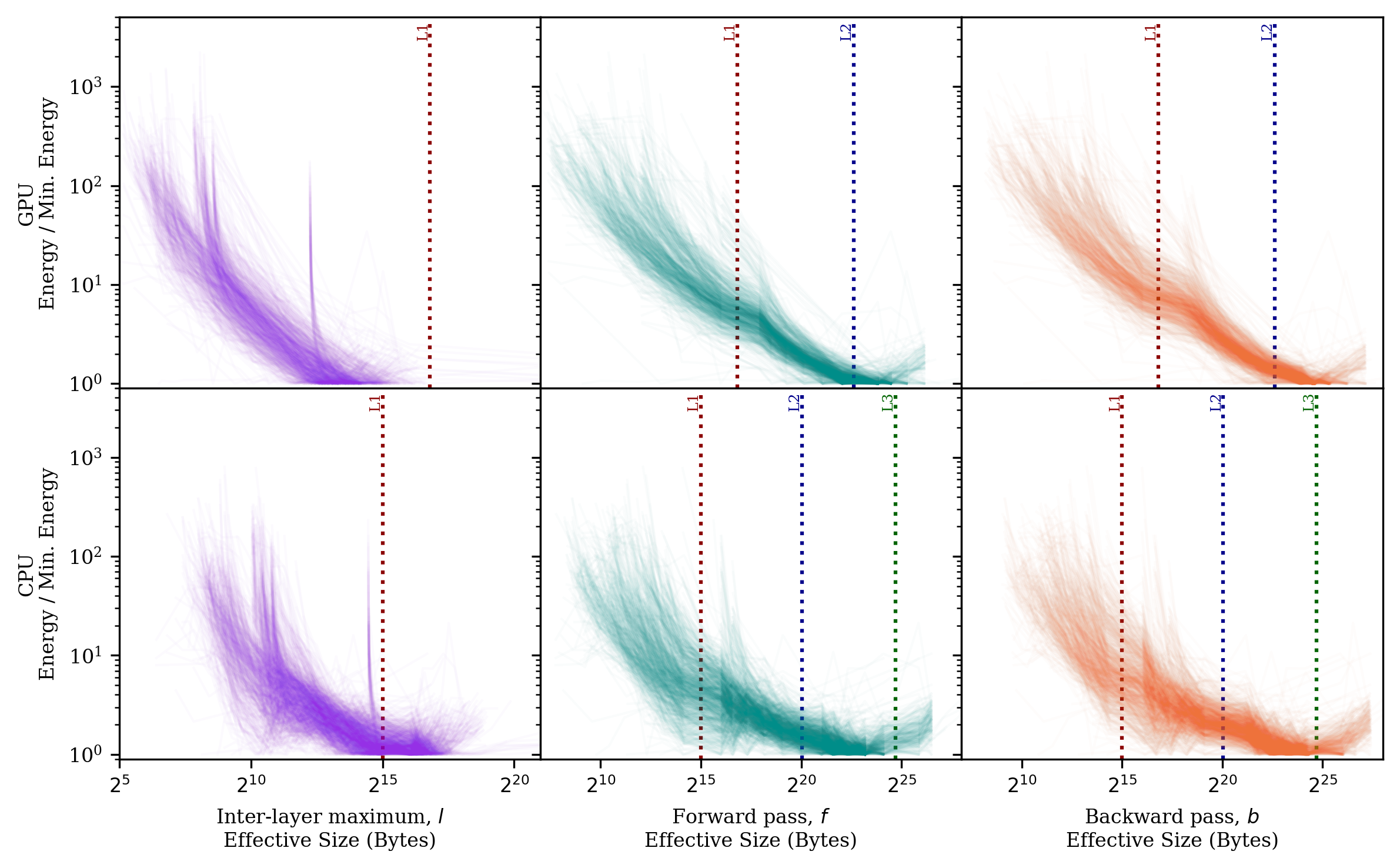}
        \caption{
        Each plot shows a selected isoloss line from each combination of dataset, shape, and depth in the BUTTER-E dataset.
        The x-axis indicates the effective size of a particular working set (from left to right: $l$, $f$, and $b$).
        The y-axis indicates the ratio of energy used to the minimum energy required to reach the isoline's test loss level.
        Vertical lines indicate cache boundaries where the working set spills into a higher memory level.
        }
        \label{fig:isolines}
    \end{subfigure}%
    \begin{subfigure}
        \centering
        \includegraphics[width=.98\linewidth]{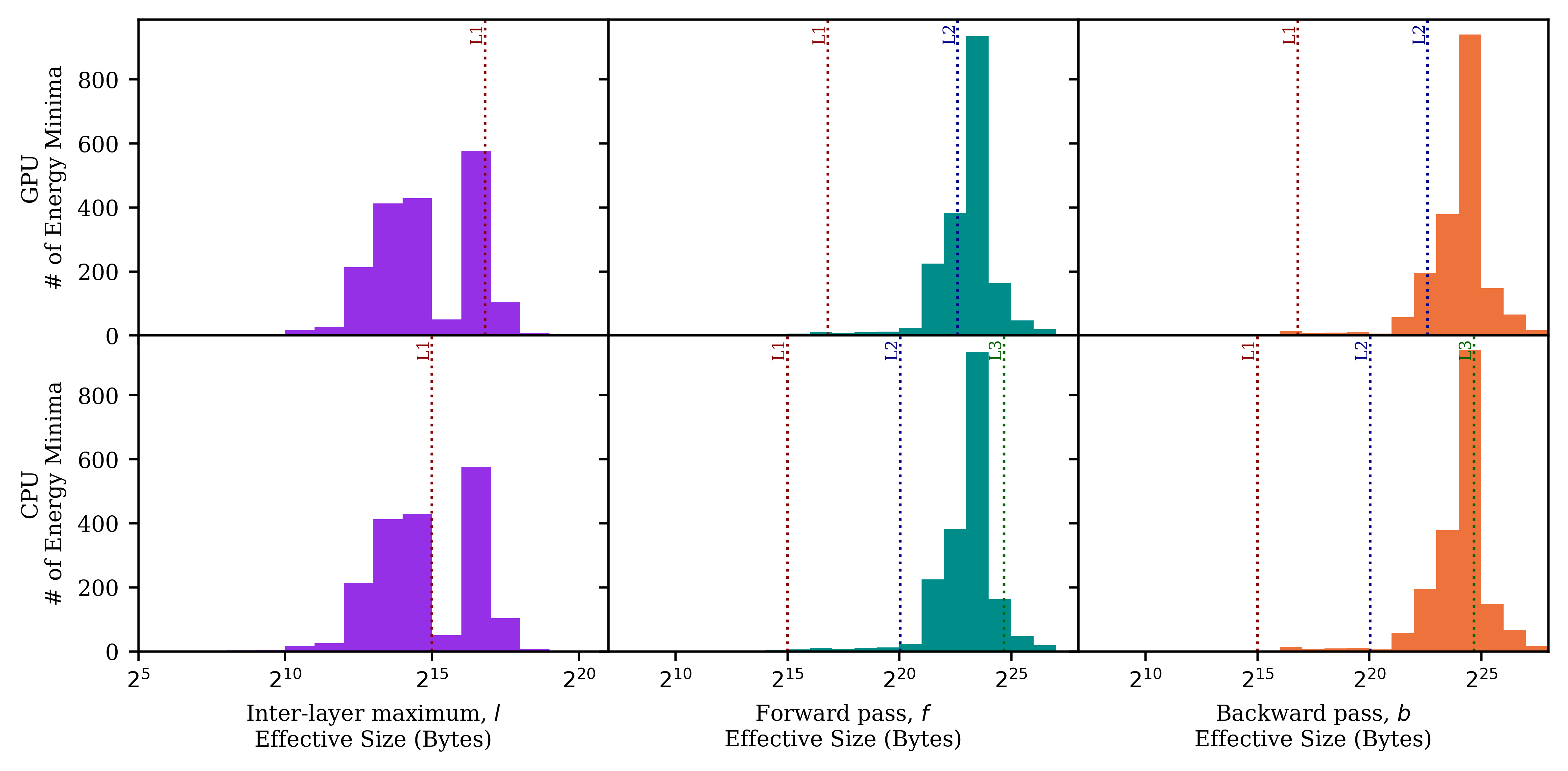}
        \caption{
            Histograms of isoline minima from the plots in Figure \ref{fig:isolines}.
            The most energy efficient working set size is typically just past the point that it spills into a higher memory level.
        }
        \label{fig:minimum_histograms}
    \end{subfigure}%
\end{figure}
As discussed in Section \ref{sec:aggregate_trends}, energy per datum increases non-linearly as NTP increases, however, energy consumed is only half of the picture: we must also consider the energy required to reach a given loss level.
As the expressive capacity of a network generally increases with NTP and therefore FLOPS, one might think that the energy required to reach a given loss level universally rises as NTP increases and that the most efficient NTP is nearly the smallest NTP capable of achieving a given loss level.

However, two facts contradict these beliefs.
First, in the BUTTER dataset the number of epochs (and therefore training data) required to reach any loss level decreases mildly as NTP increases.
That is to say that as NTP increases so does data efficiency, creating a tension between choosing a small network with lower FLOPs per datum and a large network with a higher data efficiency.
Previous work suggests that the inverse cube root of the test-loss minimizing epoch has an approximately affine relationship to the logarithm of the number of parameters,~\cite{butter_publication}:
\begin{equation}
    {(\textrm{minimizing epoch})}^{-1/3} \propto \log(\textrm{NTP}) + c
\end{equation}
Second, we observed in the preceding sections that energy per datum increases non-linearly as key working sets migrate to higher memory levels (see Figure \ref{fig:aggregate_trends}), suggesting that the balance between increasing data efficiency and increasing energy per datum is mediated by cache and working set sizes.
So, what is the most energy efficient NTP?

Figure \ref{fig:sample_isolines} plots the interpolated number of FLOPs and measured energy required to reach various loss levels over a selection of four dataset, network shape, and depth combinations.
We used linear interpolation in log-log space to generate these plots.
Qualitatively, FLOPs and actual measured energy required to achieve various test loss levels tell very different stories.
The tension between NTP and data-efficiency can be seen in all but the 201\_pol cases where the most FLOPs-efficient NTP is somewhat larger than the minimum NTP to achieve each loss level.
This is the point at which the increased data efficiency of larger NTPs is in balance with the increasing number of FLOPs required by larger NTPs.
However, using empirical energy measurements, we observe that energy used to reach a target loss level \textit{decreases} as NTP increases far beyond the point at which the number of FLOPs is minimized.
In fact, the energy used to reach a target loss level appears to reach a minimum at or just beyond the NTP that causes one of the key working sets (usually $f$ or $b$ in our dataset) to spill into a higher memory level.
\textbf{FLOPs does not appear to play a significant role in energy efficiency}.

Do these trends generalize to other hyperparameter combinations in the dataset?
In fact, they do.
Figure \ref{fig:isolines} plots a representative isoloss line from every hyperparameter combination across the entire BUTTER-E dataset, and plot these with respect to three working set sizes.
Figure \ref{fig:minimum_histograms} plots the most energy efficient working set size of each isoline in Figure \ref{fig:isolines}.
Here, we observe that across virtually all hyperparameter combinations, the energy required to reach a given loss level decreases as NTP increases until reaching a minimum at or just past where a key working set spills into a higher memory level, beyond which energy required once again increases.
In other words, the energy used to reach any target loss level in this dataset virtually always reaches a minimum within a factor of 2x of the NTP that causes one of the key working sets (usually $f$ or $b$ in our dataset) to spill into a higher memory level.

With these observations, we conclude that \textbf{the most energy-efficient network is one sized such that a key working set equals or barely exceeds a key cache size}.
In fact, in these experiments, for any dataset, shape, depth, and target test loss, the NTP that consumes the least energy to achieve any target loss level is virtually always one where the backward pass working set completely fills or just slightly spills from L2 cache.
Therefore a simple rule of thumb is that an energy efficient NTP is one that places the backward pass working set within a factor of two of a cache size.
With multiple cache levels, the most energy efficient NTP could lay near any of the cache boundaries.
For example, a good candidate for the most energy efficient network on a two cache level system, such as our V100 GPUs, would be one where $s_b$ or $s_f \approx |\textrm{L2}|$.
Simpler tasks might find their most efficient size be one where $s_b$ or $s_f \approx |\textrm{L1}|$.
More complex tasks and lower loss levels that require $s_f \gg |\textrm{L2}|$ to achieve feasibility will likely find energy efficient sizes based instead on the inter-layer working set size where $\max_{l\in L} s_l \approx |\textrm{L1}|$ or $|\textrm{L2}|$.

\section{Conclusions}
\label{sec:conclusion}

Grappling concretely with the energy implications of system design decisions will only become more critical as AI models continue an unmitigated explosion in complexity and energy costs.
This study aims to be a starting point for empirical, concrete measurement and analysis of algorithm energy costs.
Much more work lays ahead: similar studies must examine the many key deep learning architectures including large language models (LLMs), convolutional neural networks (CNNs), graph neural networks (GNNs).
And, we must use the results to optimize both our software and hardware for energy efficient execution of AI tasks.

From this study, we can point to several concrete action items:
\begin{enumerate}
    \item \textbf{Smaller networks do not always consume the least amount of energy during training.}
    When sizing a network, choose sizes that correspond to system caches. 
    For example, MLP networks with NTPs between 1x and 2x of a cache size are likely the most energy-efficient.
    \item \textbf{Beware of wide layers, particularly those with large input sizes.}
    The inter-layer working set is most efficient when it fits in low-level cache, and more so when it does not force the forward or backward pass working sets into a higher cache level.
    Avoid large inter-layer working sets that do not fit in cache, such as those created by wide layers with many inputs.
    While not measured in this work, we speculate that the inter-layer working set plays an even greater role in CNN energy consumption: large matrix sizes passed between CNN layers can easily spill into system memory.
    \item \textbf{Cache-aware deep learning}: develop deep learning approaches that decompose large models into cache-sized computational units.
    For example, an ensemble of cache-sized expert models may be substantially more energy efficient than one large model with the same total NTP.
    Pruning methods that produce cache-aware sparsity could have super-linear energy efficiency benefits.
    \item \textbf{Distribute working sets.} 
    Develop algorithms that distribute and pipeline neural network processing such that each processing unit only accesses a cache-sized portion of the model parameters.
    \item \textbf{Larger caches, please.} 
    Develop hardware capable of holding larger parameter counts in cache, even if the cache is slower or more energetically expensive, may have substantial benefits for energy efficiency.
    \item  \textbf{More caches, please.} 
    Develop hardware capable of efficiently partitioning parameter and inter-layer working sets among many computing units and caches.
    The more that the forward and backward pass working sets can be distributed, and the more caching levels mitigate energy costs, the larger a model can be and still see efficiency benefits.
    \item \textbf{Reduce idle time.}
    Codesign algorithm-informed prefetching and buffering techniques to reduce the energy cost when working sets spill out of cache.
    When memory must be accessed, doing so with as few stalls as possible will eliminate excess energy consumption due to idle power drawn while waiting for bits to arrive from memory.
\end{enumerate}
In summary, our empirical analysis sheds light on the nuanced energy consumption patterns and trade-offs of deep neural networks and underscores the importance of considering the interactions between network and hardware topologies when developing AI systems.
By integrating the insights gleaned from our study into both algorithmic and hardware designs, we can pave the way for energy-efficient AI systems that do not compromise performance.
This work, therefore, represents a crucial step toward harmonizing the rapid growth of AI capabilities with the pressing need to mitigate computing's rapidly escalating energy consumption.


\newpage

\vskip 0.2in
\bibliography{bibliography}

\begin{thebibliography}{35}
\providecommand{\natexlab}[1]{#1}
\providecommand{\url}[1]{\texttt{#1}}
\expandafter\ifx\csname urlstyle\endcsname\relax
  \providecommand{\doi}[1]{doi: #1}\else
  \providecommand{\doi}{doi: \begingroup \urlstyle{rm}\Url}\fi

\bibitem[Agency(2024)]{eGRID_data_explorer_2024}
U.~S. Environmental~Protection Agency.
\newblock Emissions \& {G}eneration {R}esource {I}ntegrated {D}atabase (e{GRID}) {D}ata {E}xplorer.
\newblock \url{https://www.epa.gov/egrid/data-explorer}, 2024.
\newblock Updated: 2024-01-30.

\bibitem[Andrae(2020)]{andrae_new_2020}
Anders S~G Andrae.
\newblock New perspectives on internet electricity use in 2030.
\newblock 2020.

\bibitem[Bakhtiarifard et~al.(2023)Bakhtiarifard, Igel, and Selvan]{bakhtiarifard_ec-nas_2023}
Pedram Bakhtiarifard, Christian Igel, and Raghavendra Selvan.
\newblock {EC}-{NAS}: {Energy} {Consumption} {Aware} {Tabular} {Benchmarks} for {Neural} {Architecture} {Search}, May 2023.
\newblock URL \url{http://arxiv.org/abs/2210.06015}.
\newblock arXiv:2210.06015 [cs, stat].

\bibitem[Brownlee et~al.(2021)Brownlee, Adair, Haraldsson, and Jabbo]{brownlee_exploring_2021}
Alexander~E.I Brownlee, Jason Adair, Saemundur~O. Haraldsson, and John Jabbo.
\newblock Exploring the {Accuracy} – {Energy} {Trade}-off in {Machine} {Learning}.
\newblock In \emph{2021 {IEEE}/{ACM} {International} {Workshop} on {Genetic} {Improvement} ({GI})}, pages 11--18, Madrid, Spain, May 2021. IEEE.
\newblock ISBN 978-1-66544-466-8.
\newblock \doi{10.1109/GI52543.2021.00011}.
\newblock URL \url{https://ieeexplore.ieee.org/document/9474356/}.

\bibitem[Cai et~al.(2017)Cai, Juan, Stamoulis, and Marculescu]{cai2017neuralpower}
Ermao Cai, Da-Cheng Juan, Dimitrios Stamoulis, and Diana Marculescu.
\newblock Neuralpower: Predict and deploy energy-efficient convolutional neural networks.
\newblock In \emph{Asian Conference on Machine Learning}, pages 622--637. PMLR, 2017.

\bibitem[Danowitz et~al.(2012)Danowitz, Kelley, Mao, Stevenson, and Horowitz]{danowitz2012cpu}
Andrew Danowitz, Kyle Kelley, James Mao, John~P Stevenson, and Mark Horowitz.
\newblock Cpu db: Recording microprocessor history: With this open database, you can mine microprocessor trends over the past 40 years.
\newblock \emph{Queue}, 10\penalty0 (4):\penalty0 10--27, 2012.

\bibitem[Deng(2012)]{deng2012mnist}
Li~Deng.
\newblock The mnist database of handwritten digit images for machine learning research.
\newblock \emph{IEEE Signal Processing Magazine}, 29\penalty0 (6):\penalty0 141--142, 2012.

\bibitem[Dennard et~al.(1974)Dennard, Gaensslen, Yu, Rideout, Bassous, and LeBlanc]{dennard1974design}
Robert~H Dennard, Fritz~H Gaensslen, Hwa-Nien Yu, V~Leo Rideout, Ernest Bassous, and Andre~R LeBlanc.
\newblock Design of ion-implanted mosfet's with very small physical dimensions.
\newblock \emph{IEEE Journal of solid-state circuits}, 9\penalty0 (5):\penalty0 256--268, 1974.

\bibitem[Dennard et~al.(1999)Dennard, Gaensslen, Yu, Rideout, Bassous, and LeBlanc]{Dennard1999}
Robert~H. Dennard, Fritz~H. Gaensslen, Hwa-Nien Yu, V.~Leo Rideout, Ernest Bassous, and Andre~R. LeBlanc.
\newblock Design of ion-implanted mosfet's with very small physical dimensions.
\newblock \emph{Proceedings of the IEEE}, 87\penalty0 (4):\penalty0 668--678, 1999.

\bibitem[Desislavov et~al.(2021)Desislavov, Mart{\'\i}nez-Plumed, and Hern{\'a}ndez-Orallo]{desislavov2021compute}
Radosvet Desislavov, Fernando Mart{\'\i}nez-Plumed, and Jos{\'e} Hern{\'a}ndez-Orallo.
\newblock Compute and energy consumption trends in deep learning inference.
\newblock \emph{arXiv preprint arXiv:2109.05472}, 2021.

\bibitem[EpochAI(2023)]{epochMachineLearningData2023}
EpochAI.
\newblock Parameter, compute and data trends in machine learning.
\newblock https://epochai.org/data/pcd, 2023.
\newblock Updated: 12/31/23.

\bibitem[Garc{\'\i}a-Mart{\'\i}n et~al.(2019)Garc{\'\i}a-Mart{\'\i}n, Rodrigues, Riley, and Grahn]{garcia2019estimation}
Eva Garc{\'\i}a-Mart{\'\i}n, Crefeda~Faviola Rodrigues, Graham Riley, and H{\aa}kan Grahn.
\newblock Estimation of energy consumption in machine learning.
\newblock \emph{Journal of Parallel and Distributed Computing}, 134:\penalty0 75--88, 2019.

\bibitem[Jones et~al.(2018)]{jones2018stop}
Nicola Jones et~al.
\newblock How to stop data centres from gobbling up the world’s electricity.
\newblock \emph{Nature}, 561\penalty0 (7722):\penalty0 163--166, 2018.

\bibitem[Kaviani and Sohn(2020)]{kaviani2020influence}
Sara Kaviani and Insoo Sohn.
\newblock Influence of random topology in artificial neural networks: A survey.
\newblock \emph{ICT Express}, 6\penalty0 (2):\penalty0 145--150, 2020.

\bibitem[Kingma and Ba(2015)]{KingBa15}
Diederik Kingma and Jimmy Ba.
\newblock Adam: A method for stochastic optimization.
\newblock In \emph{International Conference on Learning Representations (ICLR)}, San Diega, CA, USA, 2015.

\bibitem[Koomey et~al.(2011)Koomey, Berard, Sanchez, and Wong]{Koomey2011}
Jonathan~G. Koomey, Stephen Berard, Marla Sanchez, and Henry Wong.
\newblock Implications of historical trends in the electrical efficiency of computing.
\newblock \emph{IEEE Annals of the History of Computing}, 33\penalty0 (3):\penalty0 46--54, 2011.

\bibitem[Li et~al.(2022)Li, Tsourdos, and Guo]{li_transistor_2022}
Chen Li, Antonios Tsourdos, and Weisi Guo.
\newblock A {Transistor} {Operations} {Model} for {Deep} {Learning} {Energy} {Consumption} {Scaling}, May 2022.
\newblock URL \url{http://arxiv.org/abs/2205.15062}.
\newblock Number: arXiv:2205.15062 arXiv:2205.15062 [cs].

\bibitem[Liu et~al.(2021)]{liu2021freetickets}
Shiwei Liu et~al.
\newblock Freetickets: Accurate, robust and efficient deep ensemble by training with dynamic sparsity.
\newblock In \emph{Sparsity in Neural Networks: Advancing Understanding and Practice 2021}, 2021.

\bibitem[Markov(2014)]{Markov2014}
Igor~L. Markov.
\newblock Limits on fundamental limits to computation.
\newblock \emph{Nature}, 512\penalty0 (7513):\penalty0 147--154, 2014.

\bibitem[Moon et~al.(2019)Moon, Park, Rho, and Hwang]{moon2019comparative}
Jihoon Moon, Sungwoo Park, Seungmin Rho, and Eenjun Hwang.
\newblock A comparative analysis of artificial neural network architectures for building energy consumption forecasting.
\newblock \emph{International Journal of Distributed Sensor Networks}, 15\penalty0 (9):\penalty0 1550147719877616, 2019.

\bibitem[Moore(1965)]{Moore1965}
Gordon~E. Moore.
\newblock Cramming more components onto integrated circuits.
\newblock \emph{Electronics}, 38\penalty0 (8):\penalty0 114--117, 1965.

\bibitem[Olson et~al.(2017)Olson, La~Cava, Orzechowski, Urbanowicz, and Moore]{Olson2017PMLB}
Randal~S. Olson, William La~Cava, Patryk Orzechowski, Ryan~J. Urbanowicz, and Jason~H. Moore.
\newblock Pmlb: a large benchmark suite for machine learning evaluation and comparison.
\newblock \emph{BioData Mining}, 10\penalty0 (1):\penalty0 36, Dec 2017.
\newblock ISSN 1756-0381.
\newblock \doi{10.1186/s13040-017-0154-4}.
\newblock URL \url{https://doi.org/10.1186/s13040-017-0154-4}.

\bibitem[Parcollet and Ravanelli(2021)]{parcollet_energy_2021}
Titouan Parcollet and Mirco Ravanelli.
\newblock The {Energy} and {Carbon} {Footprint} of {Training} {End}-to-{End} {Speech} {Recognizers}.
\newblock In \emph{Interspeech 2021}, pages 4583--4587. ISCA, August 2021.
\newblock \doi{10.21437/Interspeech.2021-456}.
\newblock URL \url{https://www.isca-speech.org/archive/interspeech_2021/parcollet21_interspeech.html}.

\bibitem[Rodrigues et~al.(2018)Rodrigues, Riley, and Luj{\'a}n]{rodrigues2018synergy}
Crefeda~Faviola Rodrigues, Graham Riley, and Mikel Luj{\'a}n.
\newblock Synergy: An energy measurement and prediction framework for convolutional neural networks on jetson tx1.
\newblock In \emph{Proceedings of the International Conference on Parallel and Distributed Processing Techniques and Applications (PDPTA)}, pages 375--382. The Steering Committee of The World Congress in Computer Science, Computer~…, 2018.

\bibitem[Rolnick et~al.(2019)Rolnick, Donti, Kaack, Kochanski, Lacoste, Sankaran, Ross, Milojevic-Dupont, Jaques, Waldman-Brown, Luccioni, Maharaj, Sherwin, Mukkavilli, Kording, Gomes, Ng, Hassabis, Platt, and Bengio]{Rolnick2019TacklingCC}
David Rolnick, Priya~L. Donti, Lynn~H. Kaack, Kelly Kochanski, Alexandre Lacoste, Kris Sankaran, Andrew~S. Ross, Nikola Milojevic-Dupont, Natasha Jaques, Anna Waldman-Brown, Alexandra Luccioni, Tegan Maharaj, Evan~D. Sherwin, S.~K. Mukkavilli, Konrad~P. Kording, Carla Gomes, Andrew~Y. Ng, Demis Hassabis, John~C. Platt, and Yoshua Bengio.
\newblock Tackling climate change with machine learning.
\newblock \emph{arXiv}, arXiv:1906.05433, 2019.

\bibitem[Romano et~al.(2021)Romano, Le, La~Cava, Gregg, Goldberg, Chakraborty, Ray, Himmelstein, Fu, and Moore]{romano2021pmlb}
Joseph~D Romano, Trang~T Le, William La~Cava, John~T Gregg, Daniel~J Goldberg, Praneel Chakraborty, Natasha~L Ray, Daniel Himmelstein, Weixuan Fu, and Jason~H Moore.
\newblock Pmlb v1.0: an open source dataset collection for benchmarking machine learning methods.
\newblock \emph{arXiv preprint arXiv:2012.00058v2}, 2021.

\bibitem[Schwartz et~al.(2020)Schwartz, Dodge, Smith, and Etzioni]{schwartz2020green}
Roy Schwartz, Jesse Dodge, Noah~A Smith, and Oren Etzioni.
\newblock Green ai.
\newblock \emph{Communications of the ACM}, 63\penalty0 (12):\penalty0 54--63, 2020.

\bibitem[Statista(2023)]{statista_2023}
Statista.
\newblock What is the average annual power usage effectiveness ({PUE}) for your largest data center?
\newblock \url{https://www.statista.com/statistics/1229367/data-center-average-annual-pue-worldwide/}, 2023.

\bibitem[Strubell et~al.(2019)Strubell, Ganesh, and McCallum]{Strubell2019}
Emma Strubell, Ananya Ganesh, and Andrew McCallum.
\newblock Energy and policy considerations for deep learning in nlp.
\newblock \emph{arXiv}, arXiv:1906.02243, 2019.

\bibitem[{TOP500 Project}(2023)]{Top500}
{TOP500 Project}.
\newblock {TOP500 Supercomputer Sites}.
\newblock \url{https://www.top500.org/}, 2023.
\newblock Accessed: 2024-02-01.

\bibitem[Tripp et~al.(2022{\natexlab{a}})Tripp, Perr-Sauer, Hayne, and Lunacek]{butter_oedi_osti_dataset}
Charles Tripp, Jordan Perr-Sauer, Lucas Hayne, and Monte Lunacek.
\newblock Butter - empirical deep learning dataset.
\newblock 5 2022{\natexlab{a}}.
\newblock \doi{10.25984/1872441}.

\bibitem[Tripp et~al.(2022{\natexlab{b}})Tripp, Perr-Sauer, Hayne, and Lunacek]{butter_publication}
Charles~Edison Tripp, Jordan Perr-Sauer, Lucas Hayne, and Monte Lunacek.
\newblock An empirical deep dive into deep learning's driving dynamics.
\newblock \emph{arXiv preprint arXiv:2207.12547}, 2022{\natexlab{b}}.

\bibitem[Xu et~al.(2023)Xu, Martínez-Fernández, Martinez, and Franch]{xu2023energy}
Yinlena Xu, Silverio Martínez-Fernández, Matias Martinez, and Xavier Franch.
\newblock Energy efficiency of training neural network architectures: An empirical study, 2023.

\bibitem[Yang et~al.(2017)Yang, Chen, Emer, and Sze]{yang_method_2017}
Tien-Ju Yang, Yu-Hsin Chen, Joel Emer, and Vivienne Sze.
\newblock A method to estimate the energy consumption of deep neural networks.
\newblock In \emph{2017 51st {Asilomar} {Conference} on {Signals}, {Systems}, and {Computers}}, pages 1916--1920, October 2017.
\newblock \doi{10.1109/ACSSC.2017.8335698}.
\newblock ISSN: 2576-2303.

\bibitem[Ying et~al.(2019)Ying, Klein, Christiansen, Real, Murphy, and Hutter]{ying2019bench}
Chris Ying, Aaron Klein, Eric Christiansen, Esteban Real, Kevin Murphy, and Frank Hutter.
\newblock Nas-bench-101: Towards reproducible neural architecture search.
\newblock In \emph{International conference on machine learning}, pages 7105--7114. PMLR, 2019.

\end{thebibliography}

\end{document}